\documentclass{article}

\usepackage{graphicx} 
\usepackage{subcaption}

\usepackage{natbib}

\usepackage{algorithm}
\usepackage{algorithmic}
\usepackage{hyperref}
\usepackage{array}
\usepackage[table]{xcolor}
\usepackage[export]{adjustbox}
\usepackage{longtable}
\usepackage{amsmath}
\usepackage{array,multirow}

\usepackage[accepted]{icml2017}

\usepackage{tikz}
\usepackage{bm}

\newcommand\numberthis{\addtocounter{equation}{1}\tag{\theequation}}

\icmltitlerunning{Towards Visual Explanations for Convolutional Neural Networks via Input Resampling}

\begin{document}

\twocolumn[
\icmltitle{Towards Visual Explanations for Convolutional Neural Networks
via Input Resampling}


\icmlsetsymbol{equal}{*}

\begin{icmlauthorlist}
\icmlauthor{Benjamin J. Lengerich}{equal,cmu}
\icmlauthor{Sandeep Konam}{equal,cmu}
\icmlauthor{Eric P. Xing}{cmu,petuum}
\icmlauthor{Stephanie Rosenthal}{cmu}
\icmlauthor{Manuela Veloso}{cmu}
\end{icmlauthorlist}

\icmlcorrespondingauthor{Benjamin J. Lengerich}{blengeri@andrew.cmu.edu}
\icmlcorrespondingauthor{Sandeep Konam}{skonam@andrew.cmu.edu}

\icmlaffiliation{cmu}{Carnegie Mellon University,
            5000 Forbes Ave., Pittsburgh, PA 15213 USA}
\icmlaffiliation{petuum}{Petuum Inc,
            2555 Smallman Street, Pittsburgh, PA 15222 USA}

\icmlkeywords{convolutional neural networks, visualization}

\vskip 0.3in
]


\printAffiliationsAndNotice{\icmlEqualContribution} 

\begin{abstract}
The predictive power of neural networks often costs model interpretability. Several techniques have been developed for explaining model outputs in terms of input features; however, it is difficult to translate such interpretations into actionable insight. Here, we propose a framework to analyze predictions in terms of the model's internal features by inspecting information flow through the network.
Given a trained network and a test image, we select neurons by two metrics, both measured over a set of images created by perturbations to the input image: (1) magnitude of the correlation between the neuron activation and the network output and (2) precision of the neuron activation. We show that the former metric selects neurons that exert large influence over the network output while the latter metric selects neurons that activate on generalizable features. By comparing the sets of neurons selected by these two metrics, our framework suggests a way to investigate the internal attention mechanisms of convolutional neural networks.
\end{abstract}

\section{Introduction}
\label{introduction}
While the high predictive performance of deep network models can engender trust for machine learning practitioners, it is not completely sufficient for deployment. Perturbation analysis has shown that deep models can be surprisingly brittle, breaking in unexpected ways for ``adversarial" data \cite{szegedy2013intriguing}. Therefore, machine learning systems that will be deployed in high-risk situations like healthcare \cite{ching2017opportunities} must be open to interrogation. Unfortunately, current methods of debugging deep neural networks are laborious and often rely on human intuition. Restricting the class of usable models to include only human-intelligible models is a potential remedy, but these models are limited by representational power.

Thus, significant recent efforts have attempted to analyze how deep neural networks make predictions. Many works have shown that individual neurons may be responsible for activating on specific patterns (e.g., for images of animals, they may find the eyes, nose, or mouth). These patterns can arise without supervision; for instance, \cite{radford2017sentiment} trained a network to predict the succeeding character in text strings and found that a single neuron was incidentally trained to activate according to the text sentiment. However, links between activation patterns and network outputs are not well studied, and in practice, it is difficult to apply insights from network analysis to improve any particular model.

In contrast to previous work, we propose to analyze a network's prediction in terms of hidden, not input, features. Given a trained network and a test image, we sample a batch of images by applying small perturbations to the test image. From this batch, we identify a small set of ``important" neurons that have activation patterns strongly correlated with the network output. Finally, we translate these neurons into important patches in the input image via a deconvolutional neural network. This explanatory framework could be applied to robot navigation in order to understand the robot's decisions and diagnose errors. Toward this end, we apply the framework to a classifier trained to identify images of people, a common task involved in robot navigation.

Source code is available at \url{https://github.com/blengerich/explainable-cnn}.

\section{Notation}
\label{sec:notation}
Each neuron $(x, y)$ in layer $l$ in a neural network takes as input a set of signal matrices $z^{l-1}_{i,j}$ from neurons at indices $(i, j)$ in the previous layer and propagates a new signal matrix $z^l_{x,y}$ to the next layer using the following equation:
\begin{equation}
\label{eq:propagation}
z^l_{x,y} = \sum_{i,j}w^l_{i,j}\phi(z_{i,j}^{l-1})+b^l_{x,y}
\end{equation}

\section{Related Work}
\label{related_work}
Previous work \cite{ribeiro2016should} proposes a general method to explain a single prediction of any complex model by fitting simple (linear) models to data near the prediction of interest. By sampling data in a small neighborhood and training a simple model to mimic the predictions of the complex model, this method generates an interpretable model for each datapoint that is locally consistent with the uninterpretable model. However, there are few theoretical guarantees on the curvature of the space of the predictions of highly non-linear models like deep neural nets, so it is unclear if such simple models can adequately explain the predictions of deep models. More fundamentally, while this method of fitting locally-consistent, linear models gives interpretations for each prediction, it does not provide straightforward methods of improving the complex model. Instead, we would like to directly analyze the neural network.

\cite{zeiler2014visualizing} used deconvolutional networks \cite{zeiler2011adaptive} to visualize patterns that maximize activation of each neuron in a deep network (e.g., facial features such as the eyes or nose, other defining components of animals such as fur patterns, or even whole objects). However, their analysis of individual neurons and their image patch patterns was accomplished largely manually, which is impractical given the size of today’s state-of-the-art networks as well as the number of images that are tested.

A number of recent works propose to interpret deep neural networks in terms of individual predictions. \cite{samek2016evaluating} presents a method based on region perturbation to evaluate ordered collections of pixels. Similarly, \cite{shrikumar2016not} propagate activation differences against ``reference" activations to determine the importance of input features. Finally, \cite{koh2017understanding} propose to interpret deep neural networks by statistical influence functions and \cite{baehrens2010explain} inspect the directions in the input space that most affect the network output. As these approaches all focus on explaining network predictions in terms of the input features, they do not provide straightforward methods of debugging and improving the training of the networks.

More recently, \cite{al2017contextual} propose to use probabilistic graphical models to encode and explain the ``context" of each prediction. By fitting the explanatory probabilistic graphical models concurrently with the deep neural network, they achieve superior inference and interpretable explanations for each prediction. However, this framework requires an \textit{a priori} decision to use contextual explanatory models which sacrifices \textit{post-hoc} model flexibility.

APPLE \cite{Konam–2017–22207} analyzes a convolutional neural network to find neurons that are ``important" to the network classification outcome, and automatically labels patches of the input image that activate these important neurons. They introduce four measures of importance used to rank the neurons. Using the notation of Eq. \ref{eq:propagation}, these baseline metrics are based on the weights and activations of each neuron $(x, y)$ in layer $l$:
\begin{itemize}
\item Activation Matrix Sum: The sum of all values in the output activation signal matrix $z$:
\begin{displaymath}
I(l,x,y) = \sum_{row,col}z^l_{x,y}[row][col]
\end{displaymath}
\item Activation Matrix Variance: The variance of all values in the output activation signal matrix $z$:
\begin{align*}
I(l,x,y) &= \sum_{row,col}(z^l_{x,y}[row][col])^2 \\
&- (\sum_{row,col}z^l_{x,y}[row][col])^2
\end{align*}
\item Weight Matrix Sum: The sum of all values in the weight matrix $w^{l+1}$:
\begin{displaymath}
I(l,x,y) = \sum_{row,col}w^{l+1}_{x,y}[row][col]
\end{displaymath}
\item Weight Matrix Variance: The variance of all values in the weight matrix $w^{l+1}$:
\begin{align*}
I(l,x,y) &= \sum_{row,col}(w^{l+1}_{x,y}[row][col])^2 \\
&- (\sum_{row,col}w^{l+1}_{x,y}[row][col])^2
\end{align*}
\end{itemize}
These metrics examine only a single datapoint. For models with high representational power, these metrics can be sensitive to noise. Here, we propose to overcome this limitation by selecting neurons that are consistently important throughout a neighborhood of input space.

\section{Our approach}
\begin{figure*}[ht]
\centering
\setlength\fboxsep{0pt}
\setlength\fboxrule{0.25pt}
\fbox{\includegraphics[height=2.9in]{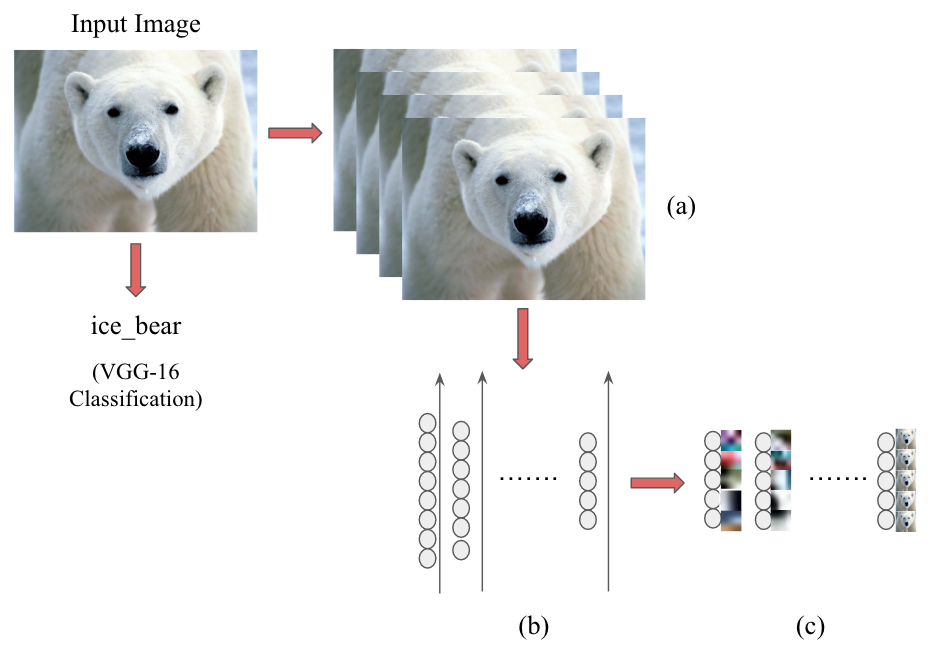}}
\caption{Pipeline of our algorithm: a) Generate a set of images by perturbations of the input image. b) Rank neurons by metrics described in Section \ref{sec:important}. c) Identify image patches corresponding to the top 5 neurons from (b).}
\label{fig:pipeline}
\end{figure*}

\label{Model}
We propose a method that combines the models of APPLE \cite{Konam–2017–22207} and LIME \cite{ribeiro2016should} to select neurons that are consistently important over a local neighborhood in the input space. First, we generate a set of input images centered at the image of interest and measure the activation of each neuron for each image in this batch. We rank the neurons by importance over this batch (Section \ref{sec:important}) and deconvolve the network to determine the visual field of each important neuron. This procedure extracts neurons and image patches that are influential in the network's output over a local area in image space. By comparing the sets of neurons selected by each metric, we can investigate the model prediction in terms of the features encoded by the hidden nodes in the network.

\subsection{Input Perturbation}
The dataset of perturbed images is generated by multiplying each pixel value in the original image by a filter of small Gaussian noise. The filter is re-created $n$ times to generate a dataset of $n$ inputs for each query image. In our experiments, we use $1+N(0,0.1)$ to generate the noise filter and repeat the process $n=50$ times. Using the notation of Eq. \ref{eq:propagation}, each $z$ thus becomes a tensor.

\subsection{Important Neurons}
\label{sec:important}
We propose two metrics: (1) Activation-Output Correlation, which approximates the influence of each neuron on the network output, and (2) Activation Precision, which estimates the generalizability of the features encoded by each neuron.

\paragraph{Activation-Output Correlation}
Magnitude of the correlation coefficient between the activation of the neuron and the final output of the network:
\begin{align*}
\label{eq:correlation}
&I(l,x,y) = \numberthis \\
&\frac{|n\sum\limits_{i=1}^n z^l_{x,y,i}o_{i} - (\sum\limits_{i=1}^nz^l_{x,y,i})(\sum\limits_{i=1}^no_i)|}{\sqrt{\Big[n\sum\limits_{i=1}^n(z^l_{x,y,i})^2 - (\sum\limits_{i=1}^nz^l_{x,y,i})^2\Big]\Big[n\sum\limits_{i=1}^n(o_{i})^2 - (\sum\limits_{i=1}^no_{i})^2\Big]}}
\end{align*}
where $o_i$ is the overall output of the network for the image with noise filter $i$.

\paragraph{Activation Precision}
\label{sec:activiation_precision}
As all of the perturbed images are generated by small perturbations to the same image, the ground truth label for each perturbed image is the same. Thus, a perfect model would output the same value for each perturbed image, i.e. $o_i = o_j = o \quad\forall i,j \in \{1,..,n\}$. After discarding neurons with activation magnitude smaller than threshold $\lambda$, we approximate the generalizability of each neuron by penalizing the variance in activations:
\begin{align*}
\label{eq:variance}
&I(l,x,y) = \numberthis\\
&\frac{1}{RC}\sum\limits_{r=1}^R\sum\limits_{c=1}^C\Big[\frac{1}{\sum\limits_{i}(z^l_{x,y,i}[r][c])^2 - (\sum\limits_{i}z^l_{x,y,i}[r][c])^2}\Big]
\end{align*}
As we show empirically (Section \ref{sec:convergence}), rankings induced by these importance metrics are similar for accurate networks and converge throughout training. This suggests possibilities for schemes to improve training efficiency by identifying hidden features that are mis-weighted by the network.

\subsection{Important Patches}
\label{sec:important_patches}
Given the ranked neurons, we select the top $N$ neurons per layer (i.e., we use the top 5 neurons). For each of the top N neurons, we are interested in identifying the image patches that they convolve. We determine the image patches by deconvolving the network using a multi-layered Deconvolutional Network (deconvnet) as in  \cite{zeiler2014visualizing}. A deconvnet can be thought of as a CNN that uses the same components (filtering, pooling) but in reverse, so instead of mapping pixels to features, a deconvnet maps features to pixels. To examine a particular neuron activation, we set all other activations in the layer to zero and pass the feature maps as input to the attached deconvnet layer. Then we successively (i) unpool, (ii) rectify and (iii) filter to reconstruct the activity in the layer beneath that gave rise to the chosen activation. This process is repeated until the input pixel space, referred to as a patch, is reached.

\section{Experiments}
\label{sec:experiments}
We trained a person vs non-person classifier using the VGG-16 \cite{simonyan2014very} network architecture on the INRIA person dataset \cite{dalal2005histograms}. The VGG-16 architecture consists of 13 convolution layers followed by 3 dense layers, with max pooling after the 2$^{nd}$, 4$^{th}$, 7$^{th}$, 10$^{th}$ and 13$^{th}$ layer. Only image patches for neurons between layers 3 and 9 were evaluated. The image patches in the first few layers (1 and 2) were too small to train a patch classifier. Furthermore, \cite{Ranzato} showed that the first layers learn stroke-detectors and Gabor-like filters. Similarly, the image patches corresponding to the last few layers (10 and above) focus on a majority of the image and thus contain too many of the object features in each patch to provide any useful insight. We ran our experiments on a laptop with 15.5 GiB RAM and 4GB GeForce GTX 960M with Intel® Core™ i7-6700HQ CPU @ 2.60GHz x 8.

\subsection{Accuracy of Patches}
\begin{figure}[ht]
\centering
\includegraphics[width=\columnwidth]{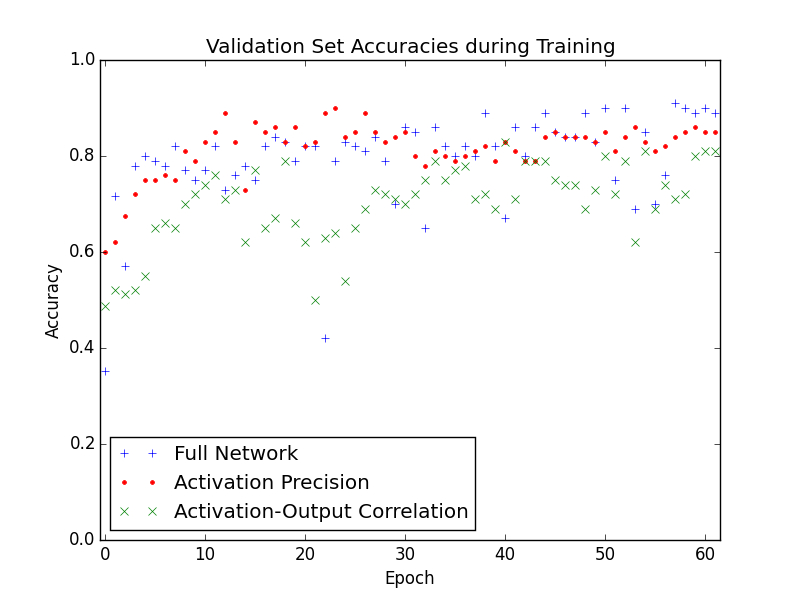}
\caption{Accuracy of a secondary classifier trained on the selected patches after each epoch of the large neural network's training.}
\label{fig:svm_accuracy}
\end{figure}

To evaluate the relevance of the patches selected by each metric, we train a secondary classifier on the collection of the top 5 patches identified as important from each image. For this classifier, we use a smaller convolutional neural net with only 64 neurons in each channel in 3 hidden layers and max-pooling in between each layer. We measure the accuracy of this classifier on the held-out validation set. As seen in Figure \ref{fig:svm_accuracy}, the patches selected by the activation precision metric outperform those selected by the activation-output correlation metric across the entire range of training epochs. We also compare these accuracy rates to those of the entire network on the full image. Intriguingly, the small CNN trained on patches selected by activation precision outperforms the full network during early stages of training. However, the performance of this method quickly levels off and the full CNN surpasses these levels by the end of training. This suggests that the activation precision metric can effectively select neurons that activate on generalizable features even if the full network is not yet tuned to these features.

\subsection{Convergence of Activation-Output Correlation and Activation Precision}
\label{sec:convergence}
\begin{figure}[ht]
\centering
\includegraphics[width=\columnwidth]{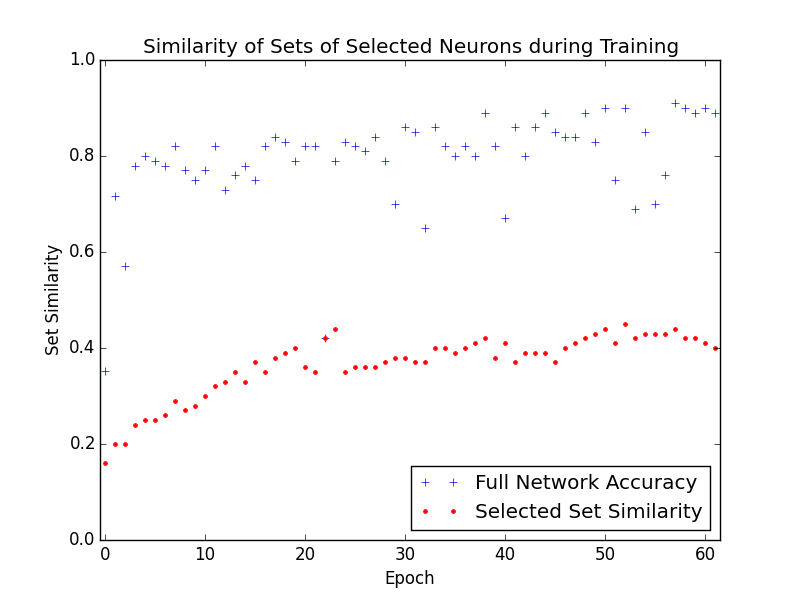}
\caption{Set similarity between neurons selected by \textit{Activation-Output Correlation} and \textit{Activation Precision} metrics increases as the classification accuracy of the network improves.}
\label{fig:similarity}
\end{figure}

As shown in Figure \ref{fig:similarity}, we also see that these two metrics converge toward selecting similar sets of neurons. We define set similarity as the Jaccard Index
\begin{align*}
    S(s_1,s_2) = \frac{|s_1\cap s_2|}{|s_1\cup s_2|}
\end{align*}
where $s_1$ and $s_2$ are the sets of neurons selected by each metric. The set similarity between the sets of neurons selected by the two metrics mirrors the convergence of the accuracy of the network.

\subsection{Weakly Supervised Localization}
\label{sec:supervised_localization}
Finally, we evaluate our algorithm's ability to select image patches that were localized on the object of interest (i.e., person in our experiments). \textit{Patch localization} is measured as the ratio of the number of selected patches containing pixels of the class to the total selected number of patches. In order to compute this ratio, we manually evaluated each image patch by comparing the patch against the original input image. In total, for each image and each selection metric, 35 important patches (top 5 patches across 7 layers) were evaluated as to whether they contained pixels representing a person (using our input image as reference).

\begin{figure*}[htp]
    \centering
    \begin{subfigure}[b]{\columnwidth}
    \centering
       \includegraphics[width=0.59\textwidth]{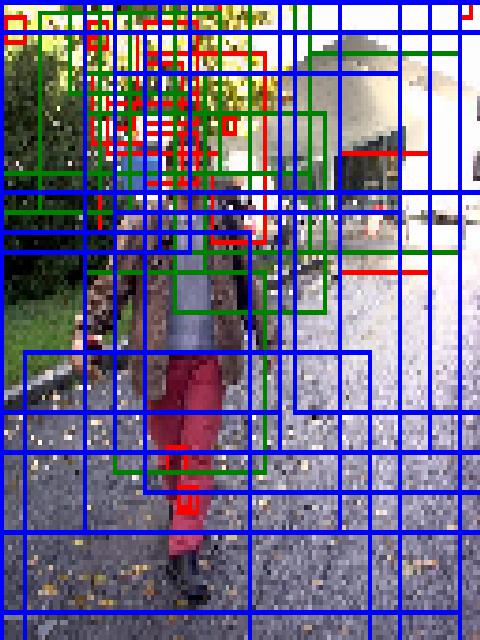}
       \caption{Activation Matrix Sum}
       \label{fig:ex_act_mat_sum}
    \end{subfigure}%
    \begin{subfigure}[b]{\columnwidth}
    \centering
       \includegraphics[width=0.59\textwidth]{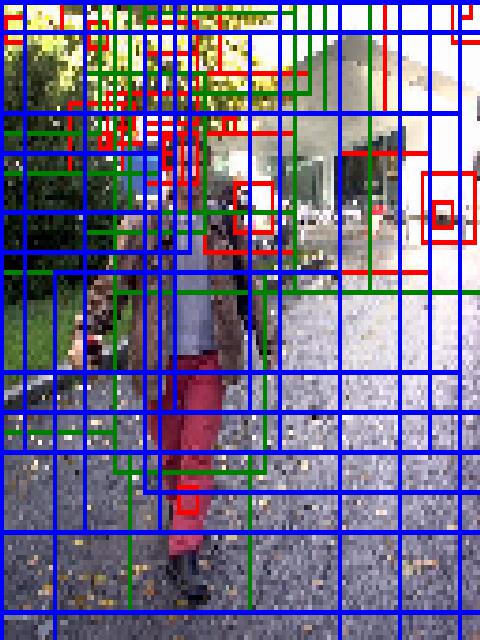}
       \caption{Activation Matrix Variance}
       \label{fig:ex_act_mat_var}
    \end{subfigure}\\
    \begin{subfigure}[b]{\columnwidth}
    \centering
       \includegraphics[width=0.59\textwidth]{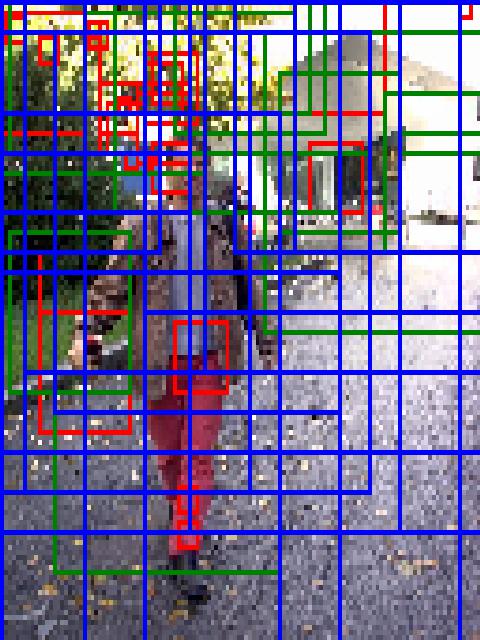}
       \caption{Weight Matrix Sum}
       \label{fig:ex_we_mat_sum}
    \end{subfigure}%
    \begin{subfigure}[b]{\columnwidth}
    \centering
       \includegraphics[width=0.59\textwidth]{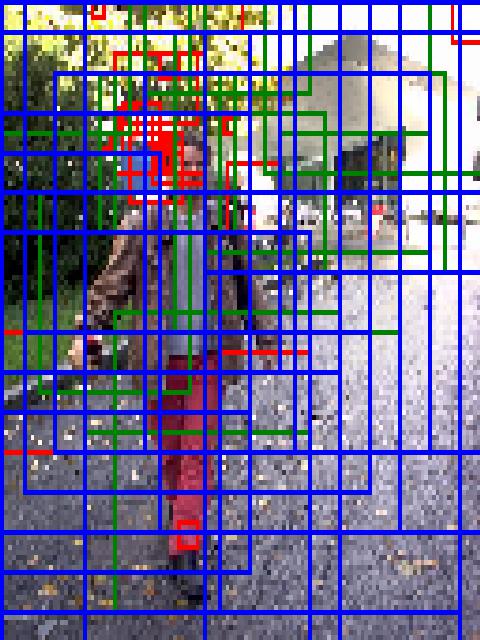}
       \caption{Weight Matrix Variance}
       \label{fig:ex_we_mat_var}
    \end{subfigure}\\
    \begin{subfigure}[b]{\columnwidth}
    \centering
       \includegraphics[width=0.59\textwidth]{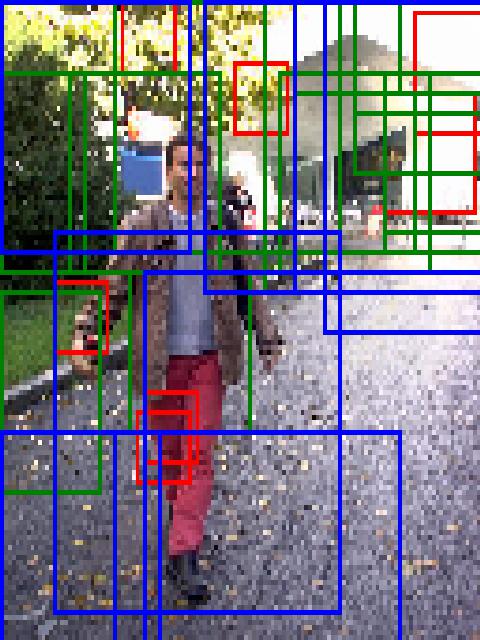}
       \caption{Activation-Output Correlation}
       \label{fig:ex_ao_cor}
    \end{subfigure}%
    \begin{subfigure}[b]{\columnwidth}
    \centering
       \includegraphics[width=0.59\textwidth]{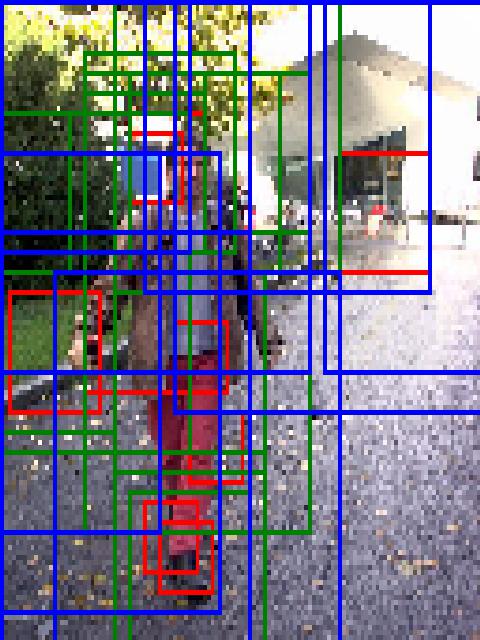}
       \caption{Activation Precision}
       \label{fig:ex_act_var}
    \end{subfigure}
    \caption{Weakly supervised localization by APPLE's selection metrics (a,b,c,d) and our proposed selection metrics (e,f). Red boxes correspond to neurons in layers 3 and 4, green layers 5 - 7, and blue boxes layers 8 and 9. Note that only a single patch in (f) does not contain the object of interest.}
    \label{fig:localization}
\end{figure*}

Figure~\ref{fig:localization} illustrates the localization abilities of our approach. Red boxes indicate patches belonging to layers 3 and 4, green boxes indicate layers 5 - 7, and blue boxes indicate layers 8 and 9. As seen from the images, most green boxes are centered around the object of interest, denoting that they are correctly identified as contributing significantly to the classification. Although we ignore patches from layers higher than 9 (as they tend to span the entire image), the patches for layers 8 and 9 (shown in blue) still cover large parts of the image. Layers 5-7 seem to capture important parts of the image that contain the person with minimal environment in the patch. Notably, on this example image, the \textit{Activation Precision} metric selects only a single patch that does not include the subject of interest.

\begin{table*}[ht]
\begin{center}
\begin{tabular}{|l|c|c|c|c|}
\hline
\multirow{2}{*}{\textbf{Selection Metric}} & \multicolumn{2}{c|}{\textbf{Top 5 Patches}} & \multicolumn{2}{c|}{\textbf{Top 20 Patches}} \\
\cline{2-5}
& \textbf{Patch Localization} & \textbf{Accuracy} & \textbf{Patch Localization} & \textbf{Accuracy}\\
\hline
Weight Matrix Sum & 0.94 & 0.607 & 0.82 & 0.703 \\
Weight Matrix Variance & 0.91 & 0.787 & 0.80 & \bf{0.948} \\
Activation Matrix Sum & 0.92 & 0.734 & 0.79 & 0.862 \\
Activation Matrix Variance & 0.91 & 0.711 & 0.80 & 0.857 \\
Activation-Output correlation & 0.93 & 0.705 & 0.85 & 0.931 \\
Activation Precision & \bf{0.96} & \bf{0.94} & \bf{0.91} & 0.901 \\
\hline
\end{tabular}
\caption{Patch Localization and Accuracy of secondary classifier on image patches selected by each metric.}
\label{tab:patches}
\end{center}
\end{table*}

Table \ref{tab:patches} shows the \textit{Patch localization} and \textit{Accuracy} values of each of the importance metrics averaged over a set of 5 test images. \textit{Accuracy} refers to the accuracy of a secondary classifier trained on the set of image patches selected by each metric. Before training, we re-size each extracted patch to be the same size (the number of images selected in each layer is kept constant across all metrics). All the metrics perform well, with \textit{Activation Precision} slightly out-performing the baseline metrics. Our results demonstrate that our proposed measure could be used to extract important neurons and patches for the purpose of understanding the reasoning behind classification in terms of the classification model.

\section{Conclusions and Future Work}
We proposed an explanatory framework to analyze the underlying reasons for a network's prediction based on information flow throughout the network. Given a trained network and a test image, our approach samples a batch of similar images, selects important neurons, and identifies corresponding patches in the input image. We present two new metrics for selecting important neurons - (1) magnitude of the correlation between neuron activation and network output, and (2) precision of neuron activation. We observe that these metrics progress toward selecting similar sets of neurons, suggesting that the correlation metric selects neurons that have large influence over the network's output, while the precision metric selects neurons neurons that activate on generalizable features. We are interested in possibilities of directly incorporating these metrics in neural network training.

\section*{Acknowledgements}
We thank Zico Kolter and Ariel Procaccia for valuable feedback, and anonymous reviewers for insightful comments.

\bibliography{submission}
\bibliographystyle{icml2017}

\end{document}